# FAST VEHICLE DETECTION AND TRACKING ON FISHEYE TRAFFIC MONITORING VIDEO USING CNN AND BOUNDING BOX PROPAGATION

*Sandy Ardianto     Hsueh-Ming Hang     Wen-Huang Cheng*

National Yang Ming Chiao Tung University

sandyardianto.ee03@nycu.edu.tw, hmhang@nctu.edu.tw, whcheng@nycu.edu.tw

## ABSTRACT

We design a fast car detection and tracking algorithm for traffic monitoring fisheye video mounted on crossroads. We use ICIP 2020 VIP Cup dataset and adopt YOLOv5 as the object detection base model. The nighttime video of this dataset is very challenging, and the detection accuracy ($AP_{50}$) of the base model is about 54%. We design a reliable car detection and tracking algorithm based on the concept of *bounding box propagation* among frames, which provides 17.9 percentage points (pp) and 6.2 pp. accuracy improvement over the base model for the nighttime and daytime videos, respectively. To speed up, the grayscale frame difference is used for the intermediate frames in a segment, which can double the processing speed.

## 1. INTRODUCTION

Vision-based traffic monitoring systems for road intersections have become increasingly popular [1] [2]. Notably, using a single fisheye camera for this job can replace multiple standard cameras. There are several commercial products available, showing the market interests. However, there are only a few public fisheye traffic monitoring video datasets available. This paper uses the ICIP 2020 VIP Cup [3] challenge dataset. Although the car identification (ID) label is not included in the ground-truth dataset, this challenge aims at detecting and tracking moving vehicles. Hence, our design goal is to detect and track moving cars. This job becomes more difficult for nighttime video because the car shape and color are often not visible, and the head or tail lights are ambiguous when used to identify cars. We like to develop a fast detection and tracking system that works well on the fisheye videos for road intersection monitoring.

We use YOLOv5 [4] as our base object detector. In Fig. 1, a fisheye image is partitioned into 3 regions: inner circle, middle ring, and outer ring. We include the car numbers and the YOLOv5 detection rate (AP50) in each region in both night and day scenes (Table 1). The cars located in the inner circle and the middle ring have much higher accuracy than that in the outer ring, where most cars are located. Also included in Table 1 are the detection rates after the use of the proposed bounding box propagation scheme for comparison. Our main contributions are as follows:

1. Design a reliable car tracking and detection algorithm based on the concept of bounding box (bbox) propagation among frames inside and between segments (a group of frames) of an image sequence. It improves the nighttime video detection accuracy by about 17.9 percentage points (pp) and the daytime video by 6.2 pp.

2. To increase the processing speed, the intermediate frames of a segment can be the grayscale frame difference. This reduces the input data size significantly and can nearly double the processing speed.

3. Because the car size is location-dependent in fisheye images, a bbox filtering algorithm is developed to eliminate the estimated bboxes exceeding the upper and lower limits. This simple checking improves accuracy by about 2 percentage points.

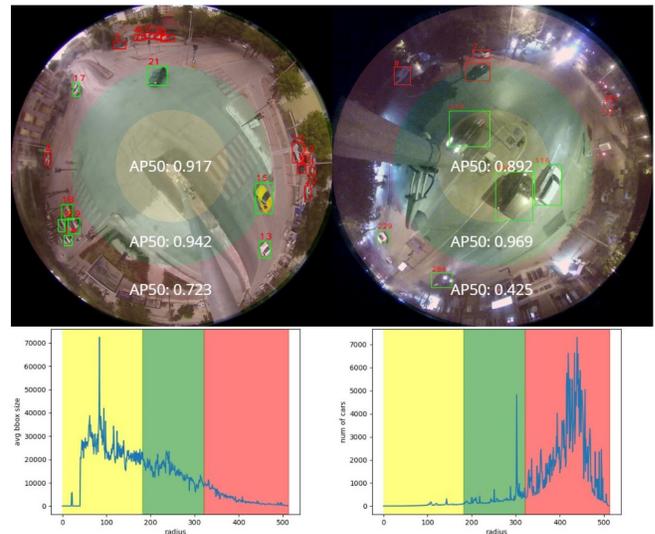

Fig. 1. ICIP 2020 VIP Cup dataset samples and statistics. (top left) day scene, (top right) night scene, (lower left) average bounding box size, (lower right) number of cars located at a given radius

In addition, we added the car ID for tracking and completed the bbox ground-truths for stationary cars manually. These additional ground truths are available at https://mcube.lab.nycu.edu.tw/icip2020vip.

## 2. RELATED WORKS

In contrast to the fisheye video dataset, the traffic monitoring videos using the standard cameras are more popular such as the City Flow dataset [5]. Sio [6] uses multiple fisheye camera tracking in the indoor environment. The existing vehicle counting or tracking systems on fisheye videos often adopt the traditional methods developed for standard cameras, such as Wang [7] which uses Kanade-Lucas-Tomasi (KLT) tracker, followed by a Hungarian algorithm. Recently, TransMOT [8] and GSDT [9] use graph neural networks to track objects. The joint detection and tracking approaches are also popular, such as FairMOT [10] and JDE [11]. Data fusion is also widely used, such as RGB-D data fusion in Jordi [12].

Since the monitoring camera is not moving, background extraction like [13] may help in improving the detection of moving objects. Our experiments indicate the simple frame difference could fulfill our detection needs. The up-sampling technique in Chen [14] may be used to restore distorted fisheye images. Tulgaç [15] uses the same dataset as ours, but it detects the cars that run outside their usual route for incident detection purposes. They rectify the fisheye distortion first before doing route detection. We did not use their approach to correct the fisheye distortion because it removes the cars in the outer ring, where most cars locate, as shown in Fig. 1. Instead, we develop a fisheye IoU tracking algorithm that uses the specific properties of fisheye video. One of them is that moving cars run along dedicated routes at nearly a constant speed. This leads to our *bounding box propagation* idea. Although this term appears in some earlier literature, their usages are quite different. For example, Lin [16] uses it inside a single image. It propagates bounding boxes from large to small objects and vice versa. Dominik [17] propagates a bounding box among frames for annotation purposes but their procedure is different from ours.

## 3. DATASET AND THEIR CHARACTERISTICS

We use ICIP 2020 VIP Cup [3] dataset for detecting and tracking the vehicles. This dataset was captured using a fisheye camera mounted on a pole near road intersections about 8 meters above the road. The image resolution is 1024x1024. There are 26 videos for training and 5 videos for testing. Each video typically has around 1000 frames captured at 15 frames per second; however, the nighttime video contains around 2000 to 4000 frames per location. Initially, the dataset has ground truth only for detection (green boxes in Fig. 1). We manually added car ID and non-moving cars (red boxes in Figure 1) for tracking purposes. There is an ambiguity in the definition of *moving cars*. If a car starts moving at frame 100, should the car have a ground-truth label before frame 100? In this dataset, this car is counted as a moving car before frame 100. Only the stationary cars for the entire sequence are not considered moving. Therefore, a stationary car detector is necessary to detect a car before it starts moving.

Based on the statistics in Fig. 1, most cars are located in the outer ring. Because our target is moving cars only, our added stationary car ground-truth is excluded in the accuracy evaluation if they do not move for the entire sequence. We divide the fisheye image into three rings based on the CKMeans clustering algorithm [18]. The first *inner circle* starts from the center with a radius of 180, the second *middle ring* with a radius from 180 to 320, and the third *outer ring* with a radius from 320 to 512. We derive the upper and lower limits of car size for each ring based on the collected data.

Another specific property was mentioned in Sec.2; namely, most moving cars run along dedicated routes at nearly a constant speed. Thus, their paths are pretty regular and tractable using overlapped bounding boxes.

## 4. PROPOSED METHOD

Based on the specific properties of the road intersection fisheye video, we designed a fast car detection and tracking algorithm. We first partition the entire sequence into segments (defined in Sec. 4.2). YOLOv5 [4] is used as the base object detector to process each frame independently. Then, the output bounding boxes exceeding the car size limits are eliminated. Next, the tracking procedure is applied to a segment of pictures, which includes a simple overlapped IoU tracking step (Sec. 4.2) and the bounding box (bbox) propagation algorithm (Sec. 4.3). The complete pipeline is shown in Fig. 2.

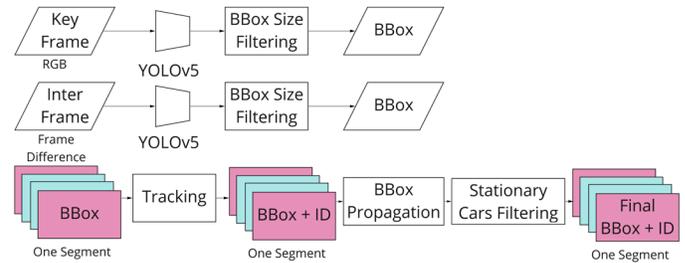

Fig. 2. Proposed Pipeline

### 4.1. Bounding Box Filtering

A simple bbox size filtering process is performed after YOLOv5 detection [4]. Since we know the distribution of car size and its location (Fig.1), we can remove a bbox with a size larger than or smaller than its usual size at its location. We collected the average bbox size for each region in Fig. 1 and added 30 pp to it to set the minimum and maximum areas (width x height of the bbox) as shown in Table 2. The square root of this area specifies the lower and upper limits of bbox height and width.

### 4.2. Segment and Tracking

A video sequence is partitioned into many segments. Each segment contains two *keyframes* (KF): the first and last frames. The other frames in a segment are called *intermediate frames* (IF). Two consecutive segments share one KF. For example, Segment 1 consists of 5 frames: frames 1 and 5 are KFs and frames 2 to 4 are IFs. Next, Segment 2 contains frames 5 to 9. Frame 5 is the KF shared by Segments 1 and 2.

To reduce computation, based on the typical car movement of our dataset, we adopt the IoU (Intersection over Union) tracking strategy [19]; that is, we assume the bboxes of a car in two neighboring frames would overlap. This simple strategy works well because the car speed passing the road intersection is usually not very fast. Using the region specification in Fig. 1, we adjust the IoU threshold based on the average car speed in each region as follows: inner circle (yellow): 0.2, middle ring (green): 0.3, and outer ring (red): 0.4. They are derived based on the dataset statistics.

### 4.3. Fast Scheme using Frame Difference

Our goal is to speed up the detection while maintaining accuracy. Our focus is moving cars. Often, the grayscale frame difference contains sufficient information to detect moving cars, which leads to a much smaller and faster detector. However, only moving cars appear in the frame difference image. Due to the moving car definition in Sec.3, we also need to detect cars before they move. We achieve this by using YOLOv5 and applying it to the original RGB image inputs (3 channels) for the KFs. We use YOLOv5 with grayscale frame difference inputs (1 channel) for the IFs to reduce computation.

### 4.4. Bounding Box Propagation

When a car is waiting for the green light, it is stationary and is not shown on the frame difference image. The KFs identify all car candidates, represented as bboxes. The first bbox propagation process copies the bboxes from KFs to IFs. However, the missed detection or false detection cases could happen in the KFs. The frame difference based IF detection may fix some errors in KFs. This leads to the second bbox propagation process from IFs to KFs. The propagation direction can be either forward or backward. Because the procedure is simple and fast, we can thus do both directions to achieve better accuracy. In the end, we add a *high confidence car* propagation algorithm to further improve detection rate.

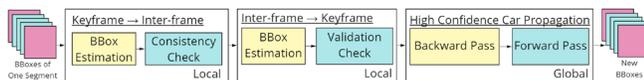

Fig. 3. Bounding box propagation pipeline
Red square: keyframe (KF); blue square: intermediate frame (IF)

A. *Keyframe to Intermediate Frame*

YOLOv5 first detects cars in all frames of a segment. Fig. 4 shows an example that three cars are detected in frame 1 (KF1), four cars are detected in frame 5 (KF5), and one car (Car 2) is detected in the three IFs. The simple IoU tracking algorithm (Sec.4.2) is first conducted and Car 2 is tracked (Fig.4(a) Left).

Next, the KF to IF propagation is performed. The bbox propagation *path* is generated by constructing the *interpolated boxes* (labeled as f2, f3, and f4) for frames 2, 3, and 4 as shown in Fig.4(b). This bbox interpolation includes two parts. We first use linear interpolation to calculate the positions (centers) of the interpolated boxes and then calculate the height and width of each interpolated box based on the heights and widths of two KF bboxes. The *path IoU* value (e.g., $IoU_{1,4}$ for Car 1 in KF1 and Car 4 in KF5) is calculated for each box pair candidate using the formula shown in Fig.4(c).

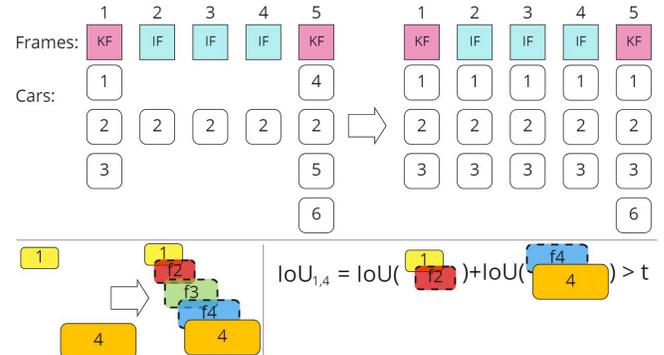

Fig. 4. KF to IF propagation example. (a) (upper) Detected cars in each frame, before and after bbox propagation, (b) (lower left) Bounding box estimation example: the interpolated boxes, f2, f3, and f4 are generated using the attributes of bboxes 1 and 4; (c) (lower right) Path IoU calculation based on the IoU values between the bboxes of KF1 and KF2 and the bboxes of their nearest IFs.

We construct all the bbox *propagation paths* from each bbox in frame 1 to all the bboxes in frame 5 and compute the IoU value associated with each path. For each bbox in frame 1, the highest IoU path is selected. If its IoU value is higher than the threshold, this path (bbox) is *valid*. Then, its ID is copied to the associated bbox in frame 5. Otherwise, that bbox is marked *invalid*. This last step is called *Consistency Check*. In the end, Fig.4(a) contains only three valid pairs (Cars 1, 2, and 3), and Car 6 is invalid at this point but it is retained on the candidate list for further processing.

B. *Intermediate Frame to Keyframe*

The example in Fig. 5 shows a candidate box A2 is detected by YOLOv5 in frame 2, box A3 in frame 3, and box A4 in frame 4. We first construct all possible bbox paths for frames 2 and 3 and evaluate their path IoU. We pick up the highest valid path to pair them. Then, we construct an *extrapolated box* A1 for frame 1 based on a valid pair A2 and A3. The center and the height and width of A1 are estimated based on the centers and the heights and widths of A2 and A3 using linear extrapolation. Then, the extrapolated box forms the input to a simple binary classifier (ResNet18) to check whether it contains a car. If so, a new car ID is assigned. This last step is called *Validation Check*.

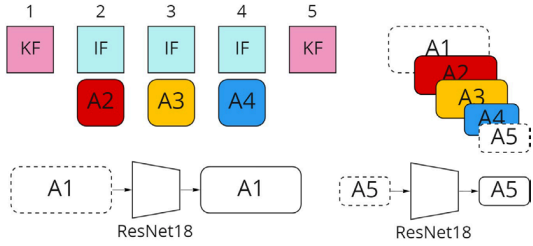

Fig. 5. IF to KF propagation example

For frame 5, the extrapolated box A5 is constructed based on A3 and A4 in a similar way. We also do a Validation Check. The binary classifier was trained using 1,000 crops of car images and 1,000 random crops from the dataset. Among them, 1,400 images (70%) are used for training, 400 (20%) for validation, and 200 (10%) for testing. The input image is resized to 128x128 for training and testing.

Table 1: Performance of the proposed scheme with bbox propagation in various regions.
The numbers in () are the percentage improvement in AP50.

| AP50 | Night | | Day | | Both | |
|---|---|---|---|---|---|---|
| | YOLOv5 | +BBox Prop | YOLOv5 | +BBox Prop | YOLOv5 | +BBox Prop |
| Inner circle | 0.892 | 0.886 | 0.917 | 0.911 | 0.864 | 0.875 |
| Middle ring | 0.969 | 0.962 | 0.942 | 0.934 | 0.965 | 0.955 |
| Outer ring | 0.425 | 0.655 | 0.723 | 0.878 | 0.695 | 0.842 |
| Overall | 0.541 | **0.720** (+17.9 pp) | 0.820 | **0.882** (+6.2 pp) | 0.772 | **0.854** (+8.2 pp) |

Table 3: Bounding box size filtering results ("After"=bboxes exceeding the size limits are removed)

| Inputs | Inner Ring | | Middle Ring | | Outer Ring | | # Cars | | | $AP_{50}$ | |
|---|---|---|---|---|---|---|---|---|---|---|---|
| | Min. Size | Max. Size | Min. Size | Max. Size | Min. Size | Max. Size | Ground Truth | Before | After | Before | After |
| Night | 113 | 162 | 92 | 132 | 53 | 76 | 19,533 | 21,423 | 20,352 | 0.541 | 0.563 |
| Day | 217 | 311 | 166 | 238 | 94 | 135 | 22,973 | 24,195 | 23,846 | 0.820 | 0.826 |
| Both | 211 | 302 | 133 | 191 | 88 | 127 | 42,506 | 45,115 | 42,702 | 0.772 | 0.781 |

### C. High Confidence Car (HCC) Propagation

Because the outer detection rate is lower, we also propagate keyframe cars with a high probability score ($\geq 0.8$), called *High Confidence Car (HCC)*, between frames. For example, if an HCC is detected in two KFs of the current segment (say, frame 97 and frame 101) and it was not detected in the previous frames, then we propagate this bbox (car) to the previous frame (frame 96) and check whether this extrapolated bbox is co-located with any of the existing bboxes. If yes, it is an *existing car* case. If not, perform a Validation Check. If a car is detected, then we propagate the car ID to this bbox and continue propagating this HCC to frame 95. If the Validate Check fails, it is a *failure* case. We continue doing the HCC propagation attempt until the existing car count or the failure count reaches a limit (=3). The above procedure is the HCC propagation *backward pass*. A similar *forward pass* is conducted for the future frames.

At the end of the bbox propagation process, we perform *stationary car filtering* by examining the centers of a valid pair of bboxes of two KFs. If the distance of their centers is less than a threshold, this car (bbox) is labeled "stationary". If its status is not changed to moving at the end of the entire sequence, it is classified as a *stationary car* and is deleted from the output list for accuracy evaluation.

## 5. EXPERIMENTAL RESULTS

In our experiments, the image is resized to 640x640 and fed into the YOLOv5 object detector. The initial model was pretrained with the COCO dataset [20]. The speed comparison is measured on a PC with an Intel i9-9900X CPU and a single NVIDIA 1080Ti GPU. Experiment settings are as follows: (or stated otherwise)

- Test data: Night video + Day video (both)
- Modality (of inputs): Keyframe: RGB, Intermediate frame: frame difference
- Segment size: 3 (KF: 2, IF: 1)
- High confidence car propagation: threshold: 0.8, and success & fail limit: 3
- Car validation network: ResNet18, input size: 128x128

### 5.1. Detection rates in 3 regions for YOLOv5

It is clear in Table 1 that the detection rates of the outer ring are much lower (than the total average), particularly, in the night videos.

### 5.2. Single Image Detection

We train YOLOv5 [4] with each input modality, namely, RGB, frame difference (FD) in RGB, and FD in grayscale.

Table 2 shows the inference results on the ICIP20 VIP testing dataset. Because FD in grayscale has only one input channel, the model size is much smaller, and computational speed (frames per second) is much faster. FD in RGB and FD in grayscale have about the same accuracy.

Table 2: Single image detection comparison ($AP_{50}$ [20])

| Input | Night | Day | Both | Speed (fps) |
|---|---|---|---|---|
| RGB | **0.541** | **0.820** | **0.772** | 32.06 |
| FD in RGB | 0.530 | 0.783 | 0.762 | 32.06 |
| FD in Grayscale | 0.524 | 0.781 | 0.759 | **94.61** |

### 5.3. Bounding box size filtering results

We apply the bbox size filtering to eliminate the estimated bbox exceeding the bbox limits in each region. The bounding box size filtering process offers about 2 pp AP50 improvement for Night video and about 1 pp for all datasets (Table 3).

### 5.4. Modality variation for bbox propagation

If we use different input modalities for our bounding box propagation schemes, the results are shown in Table 4. If every frame (including both keyframe and intermediate frame) is RGB YOLO detection, the detection accuracy is 1 percentage point higher but its speed is much slower.

### 5.5. Ablation study

We propose three bounding box propagation algorithms ((a) Keyframe to Intermediate-frame (K → I), (b) Intermediate-frame to Keyframe (I → K), and (c) High Confidence Car (HCC) propagation). Table 5 provides the performance of all possible combinations of using these three algorithms. We notice that a single propagation algorithm may not be effective but their combined use shows better performance than the individual.

Table 4. Modality variation results

| Keyframe | Inter-frame | AP50 | Speed (fps) |
|---|---|---|---|
| RGB | RGB | **0.863** | 30 |
| RGB | FD | 0.854 | 58 |
| FD | RGB | 0.783 | 46 |
| FD | FD | 0.676 | **84** |

Table 5. Evaluation results for various combinations of bbox propagation schemes

| K→I | I→K | HCC | AP50 | Speed (fps) |
|---|---|---|---|---|
|  |  |  | 0.759 | **74** |
| ● |  |  | 0.771 | 68 |
|  | ● |  | 0.728 | 69 |
|  |  | ● | 0.753 | 66 |
| ● | ● |  | 0.778 | 65 |
| ● |  | ● | 0.802 | 62 |
|  | ● | ● | 0.757 | 62 |
| ● | ● | ● | **0.854** | 58 |

### 5.6. Car Validation check network

We try different models and input sizes for the neural network used by the car Validation checking process. The results are shown in Table 6. We finally adopt ResNet50 with an input size of 128x128. It has a good detection rate although it is slower.

Table 6. Car validation network performance with different models and input sizes

| Model | Inputs | Network Accuracy | AP50 | Speed (fps) |
|---|---|---|---|---|
| ResNet18 | 32x32 | 96.8 | 0.848 | **62** |
|  | 64x64 | 97.2 | 0.853 | 60 |
|  | 128x128 | 98.9 | 0.854 | 58 |
| ResNet50 | 32x32 | 99.0 | 0.856 | 54 |
|  | 64x64 | 99.2 | 0.857 | 53 |
|  | 128x128 | **99.5** | **0.862** | 52 |

### 5.7. Parameters of the High Confidence Car (HCC) propagation

The HCC propagation algorithm has three parameters: the threshold used to select the high confidence car and the two upper limits in the stopping rule: the number of existing car attempts and the number of validation failure attempts. As shown in Table 7, a confidence threshold of around 0.8 gives the best results (Table 6). Tables 8 and 9 show that there is nearly 1 pp accuracy improvement when the validation failure limit increases from one to two. On the other hand, the existing car checking limit provides only a minor improvement, particularly when its value is higher than 2. For the upper limits of the attempts, Fig. 6 indicates a trade-off between accuracy and speed. It seems that the accuracy saturates if the attempt limit is higher than 3. Hence, we choose 3 as the upper limit for both the existing car and validation failure attempts.

Table 7. Impact of threshold values on high confidence car propagation

| Threshold | 0.5 | 0.6 | 0.7 | **0.8** | 0.9 |
|---|---|---|---|---|---|
| **AP50** | 0.684 | 0.795 | 0.837 | **0.854** | 0.853 |

Table 8. Impact of existing and failure attempt limits – AP50

| Fail \ Existing | 1 | 2 | 3 | 4 | 5 |
|---|---|---|---|---|---|
| 1 | 0.831 | 0.839 | 0.845 | 0.849 | 0.852 |
| 2 | 0.839 | 0.847 | 0.851 | 0.854 | 0.856 |
| 3 | 0.845 | 0.851 | 0.854 | 0.856 | 0.858 |
| 4 | 0.849 | 0.854 | 0.856 | 0.858 | 0.859 |
| 5 | 0.852 | 0.856 | 0.858 | 0.859 | 0.859 |

Table 9. Impact of existing and failure attempts limit – Speed (fps)

| Fail \ Existing | 1 | 2 | 3 | 4 | 5 |
|---|---|---|---|---|---|
| 1 | 64 | 63 | 62 | 61 | 59 |
| 2 | 63 | 62 | 61 | 59 | 57 |
| 3 | 62 | 61 | 59 | 57 | 55 |
| 4 | 61 | 59 | 57 | 54 | 53 |
| 5 | 59 | 57 | 55 | 53 | 51 |

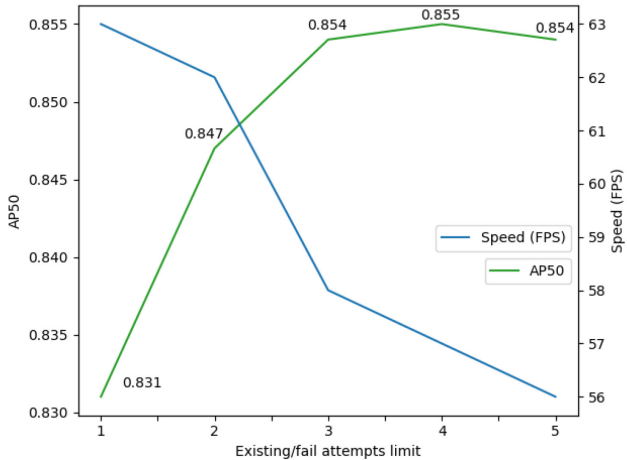

Fig. 6. High confidence car performance and speed trade-off (Same number of existing and failed attempts)

### 5.7. Bounding Box Propagation

Fig. 7 shows the detection accuracy and speed for various segment sizes. The best accuracy is achieved using segment size 3. It improves the detection accuracy (AP50) by about 17.9 pp for the night scene and 6.2 pp for the day scene compared to the single RGB input for all frames. As shown in Table 1, The improvement is more significant in the outer ring and for the night scenes. The night video has more distortion and noise compared to the daytime. Most of the cars in this dataset have white color and white headlights. At night, headlight reflection sometimes obstructs the car in front of it. The low light condition makes more grain noise than in the daytime. The detection rate of YOLOv5 in the daytime is already 0.821, and the nighttime is 0.541. There are more detection gaps between frames in the nighttime compared to daytime. Therefore, this bounding box propagation has more opportunities of filling the missing detection between frames in the night video. In addition, the detection speed improves about two times (60 fps vs. 32 fps). We can thus use grayscale frame differences to speed up the detection process.

### 5.8. Car Tracking

Our scheme assigns a car ID to each bbox and tracks the car crossing the entire image. Table 10 compares our tracking results to several well-known tracking methods. For a fair comparison, we feed the base detector outputs (center or bbox) into these tracking algorithms. However, the FairMOT [10] algorithm integrates the detection and the tracking steps together. Hence, it uses its own detection outputs. The speed listed here includes the time for the detection process. All methods in Table 10 (except ours) use RGB as the input modality. Our method uses segment size 3, with RGB for the KFs and FD for the IFs. Clearly, our scheme is the best in speed and tracking performance.

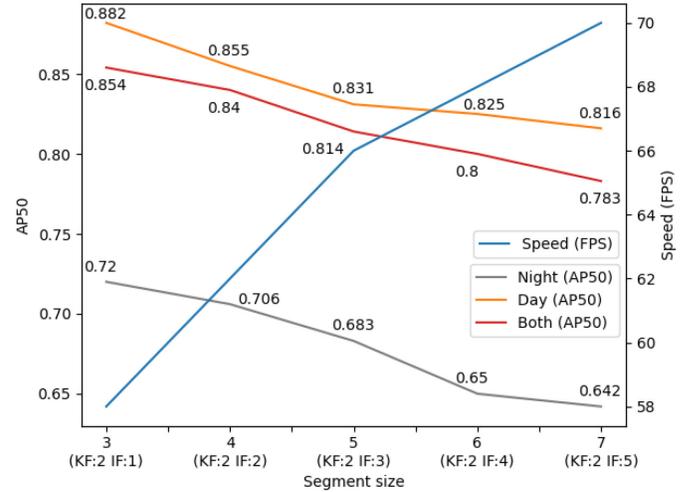

Fig. 7. Bounding box propagation performance

Table 10: Tracking comparison (MOTA and MOTP [21])

| Method | Inputs | MOTA | MOTP | Speed (fps) |
|---|---|---|---|---|
| Kalman Filter | C | 54.0 | 0.501 | 29 |
| SORT [22] | B | 56.4 | 0.535 | 26 |
| DeepSORT [23] | D | 62.2 | 0.583 | 18 |
| *FairMOT [10] | D | 76.8 | 0.789 | 13 |
| ByteTrack [24] | B | 79.2 | 0.806 | 26 |
| IoU Tracking (ours) | B | **81.0** | **0.819** | **54** |

C: Center position of the object, B: Bounding box of the object,
D: Deep learning features, *detection inputs not from our detector outputs

### 6. CONCLUSION

This paper proposes an efficient detection and tracking system that improves the detection rate on fisheye traffic monitoring video. Our design is motivated by the specific properties of the dataset. It improves the detection rate by 17.9 pp in the night scenes and 6.2 pp for the day scenes, and it increases the inference speed by nearly two times. Due to limited space, details of algorithms and experiments (including video visualization) are provided at https://mcube.lab.nycu.edu.tw/icip2020vip.

### 7. ACKNOWLEDGEMENT

This work is partially supported by the Ministry of Science and Technology, Taiwan under Grant MOST 109-2634-F-009-020.